\documentclass[hidelinks, 10pt]{article}

\usepackage[utf8]{inputenc}
\usepackage{lmodern}         
\usepackage{geometry}
\usepackage{setspace}

\usepackage{graphicx}
\usepackage{subcaption}
\graphicspath{{img/}}

\usepackage{float}
\usepackage{longtable}
\usepackage{array}
\usepackage{rotating}

\usepackage{multirow}
\usepackage{makecell}
\usepackage{ragged2e}
\usepackage{booktabs}
\usepackage{listings} 
\usepackage{amsmath, amssymb, amsfonts}
\usepackage{fancyvrb}
\usepackage{verbatim}
\usepackage{fvextra}
\usepackage{enumitem}

\usepackage{titlesec}
\usepackage{tocloft}

\usepackage{caption}
\usepackage{hyperref}

\usepackage{etoolbox}
\AtBeginEnvironment{tabular}{\fontsize{10pt}{12pt}\selectfont}
\AtBeginEnvironment{longtable}{\fontsize{10pt}{12pt}\selectfont}
\AtBeginEnvironment{table}{\fontsize{10pt}{12pt}\selectfont}

\titleformat{\section}[block]
{\centering\normalfont\fontsize{14pt}{16pt}\bfseries\selectfont}
{\thesection.}{0.5em}{}  
\titlespacing{\section}{0pt}{5pt}{0pt} 

\titleformat{\subsection}[block]
{\normalfont\fontsize{12pt}{12pt}\bfseries\selectfont}
{\thesubsection.}{0.5em}{}  
\titlespacing{\subsection}{0pt}{5pt}{0pt}

\titleformat{\subsubsection}[block]
{\normalfont\fontsize{11pt}{11pt}\bfseries\selectfont}
{\thesubsubsection.}{0.5em}{}  
\titlespacing{\subsubsection}{0pt}{5pt}{0pt}

\titleformat{\paragraph}[block]
{\normalfont\fontsize{10pt}{10pt}\bfseries\selectfont}
{\theparagraph.}{0.5em}{}
\titlespacing{\paragraph}{0pt}{5pt}{0pt}

\setlist[enumerate]{itemsep=5pt, topsep=5pt, partopsep=0pt, parsep=0pt, labelsep=0.5em, left=0pt}

\usepackage[
backend=biber,
style=numeric-comp,
sorting=none,
sortcites=true,
doi=true,        
url=true,
isbn=false
]{biblatex}

\usepackage{glossaries}
\makeglossaries
\newacronym{arf}{ARF}{Acute Respiratory Failure}
\newacronym{ards}{ARDS}{Acute Respiratory Distress Syndrome}
\newacronym{nlp}{NLP}{Natural Language Processing}
\newacronym{ehr}{EHR}{electronic health record}
\newacronym{ehrs}{EHRs}{electronic health records}
\newacronym{eicu}{eICU-CRD}{eICU Collaborative Research Database}
\newacronym{n3c}{N3C}{National COVID Cohort Collaborative}
\newacronym{llms}{LLMs}{Large Language Models}
\newacronym{llm}{LLM}{Large Language Model}
\newacronym{pasc}{PASC}{Post-Acute Sequelae of SARS-CoV-2 infection}
\newacronym{sars}{SARS-CoV-2}{SARS-CoV-2}
\newacronym{imv}{IMV}{Invasive Mechanical Ventilation}
\newacronym{nippv}{NIPPV}{Noninvasive Positive Pressure Ventilation}
\newacronym{hfni}{HFNI}{High-Flow Nasal Insufflation}
\newacronym{niv}{NIV}{Noninvasive Ventilation}
\newacronym{icu}{ICU}{Intensive Care Unit}
\newacronym{gpt}{GPT}{generative pre-trained transformer}
\newacronym{gpu}{GPU}{graphics processing unit}
\newacronym{gpus}{GPUs}{graphics processing units}
\newacronym{api}{API}{application programming interface}
\newacronym{apis}{APIs}{application programming interfaces}
\newacronym{bert}{BERT}{Bidirectional Encoder Representations from Transformers}
\newacronym{rnn}{RNN}{recurrent neural network}
\newacronym{rag}{RAG}{retrieval-augmented generation}
\newacronym{icd}{ICD-10}{International Classification of Diseases (version 10)}
\newacronym{sql}{SQL}{Structured Query Language}
\newacronym{roc}{AUROC}{area under the receiver operating characteristic}
\newacronym{prc}{AUPRC}{area under the precision-recall curve}
\newacronym{cocomo}{COCOMO}{Constructive Cost Model}
\newacronym{phi}{PHI}{Protected Health Information}
\newacronym{slms}{SLMs}{statistical language models}
\newacronym{iqr}{IQR}{Interquartile range}
\newacronym{cot}{CoT}{Chain-of-Thought}
\newacronym{pheona}{PHEONA}{Evaluation of PHEnotyping for Observational Health Data}
\newacronym{omop}{OMOP-CDM}{Observational Medical Outcomes Partnership Common Data Model}
\newacronym{framework}{SHREC}{SHifting to language model-based REal-world Computational phenotyping}
\newacronym{amia}{AMIA}{American Medical Informatics Association}
\newacronym{bleu}{BLEU}{Bilingual Evaluation Understudy}
\newacronym{rouge}{ROUGE}{Systematized Nomenclature of Medicine—Clinical Terms}

\newglossaryentry{snomed}{
	name=SNOMED CT,
	description={Systematized Nomenclature of Medicine—Clinical Terms}
}

\begin{document}
	
\begin{center}
	\textbf{\fontsize{14pt}{14pt}\selectfont 
		\acrshort{pheona}: An Evaluation Framework for \acrlong{llm}-based Approaches to Computational Phenotyping}
	
	\vspace{1.5ex}
	
	\noindent
	Sarah A. Pungitore, MS$^1$, Shashank Yadav, MS$^2$, Vignesh Subbian, PhD$^2$ \\[1ex]
	$^1$ Program in Applied Mathematics, The University of Arizona, Tucson, AZ \\
	$^2$ College of Engineering, The University of Arizona, Tucson, AZ
\end{center}

\section*{Abstract}

\textit{Computational phenotyping is essential for biomedical research but often requires significant time and resources, especially since traditional methods typically involve extensive manual data review. While machine learning and natural language processing advancements have helped, further improvements are needed. Few studies have explored using \acrfull{llms} for these tasks despite known advantages of \acrshort{llms} for text-based tasks. To facilitate further research in this area, we developed an evaluation framework, \acrfull{pheona}, that outlines context-specific considerations. We applied and demonstrated \acrshort{pheona} on concept classification, a specific task within a broader phenotyping process for \acrfull{arf} respiratory support therapies. From the sample concepts tested, we achieved high classification accuracy, suggesting the potential for \acrshort{llm}-based methods to improve computational phenotyping processes.}

\section*{Introduction}

Clinical phenotyping, the process of extracting information from health data to classify patients based on relevant characteristics, is fundamental for biomedical research\cite{hripcsak_next-generation_2013}. One such type of these processes is electronic or computational phenotyping, which involves defining cohorts algorithmically using data from \acrfull{ehrs} and other relevant information systems\cite{callahan_characterizing_2023}. The resulting phenotypes have been used for many clinical outcomes and conditions supporting a variety of downstream tasks, including recruitment for clinical trials, development of clinical decision support systems, and hospital quality reporting\cite{banda_advances_2018, tekumalla_towards_2024, shang_making_2019}. The process of developing computable phenotypes  includes constructing relevant data elements for classification and then applying an algorithm to produce the cohort(s) of interest\cite{carrell_general_2024}. For example, a computable phenotype for \acrfull{pasc} was developed by first identifying standard clinical concepts for each \acrshort{pasc}-related symptom and then applying an algorithm to identify individuals with \acrshort{pasc} based on symptom presence relative to \acrshort{sars} infection using \acrshort{ehr} data\cite{pungitore_computable_2024}.
\\
\\
\noindent
Traditionally, development of computable phenotypes requires time-intensive, manual review of data and definitions, including identification of relevant data and mapping to controlled vocabularies\cite{shang_making_2019}. From the \acrshort{pasc} example, $6{,}569$ concepts across $151$ symptoms were manually reviewed over several weeks by a clinician expert to determine relevance of each concept to \acrshort{pasc}-specific symptoms\cite{pungitore_computable_2024}. While developments in \acrfull{nlp} and machine learning methods could improve processing and categorization of clinical data, these methods also add additional steps such as model training and hyperparameter tuning which can be similarly time and resource-intensive\cite{banda_advances_2018, shang_making_2019}. Thus, there is a need for substantial improvements to traditional computational phenotyping processes, particularly for tasks that require manual review or intervention. As a relatively new addition to biomedical research, \acrfull{llms} introduce a brand new set of text analysis and generation capabilities that could supplement or replace manual review of clinical data elements\cite{raiaan_review_2024}. Furthermore, \acrshort{llms} can be trained on new tasks through prompt engineering alone, which is a more accessible and portable method of model adaptation compared to traditional model retraining or domain adaptation methods\cite{liu_pre-train_2023}. Therefore, we hypothesize \acrshort{llms} can transform computational phenotyping, in particular through acceleration and enrichment of manual review through incorporation of \acrshort{llm}-supported processes.
\\
\\
\noindent
To our knowledge, there have been no studies that have explored the application of \acrshort{llms} to computational phenotyping tasks. Therefore, this study makes the following contributions:

\begin{enumerate}
    \item We developed \acrshort{pheona} (\acrlong{pheona}), a framework that outlines an evaluation mechanism for \acrshort{llm}-based computational phenotyping.
    
    \item We provided an example usage of all components of \acrshort{pheona} to promote further research and advancements in computational phenotyping processes.

    \item We demonstrated a successful application of \acrshort{llm}-based methods to concept classification, a step within a broader phenotyping process that was previously performed solely via manual review.
\end{enumerate}

\section*{Methods}

\begin{figure}[H]
  \centering
    \includegraphics[width=0.40\textwidth]{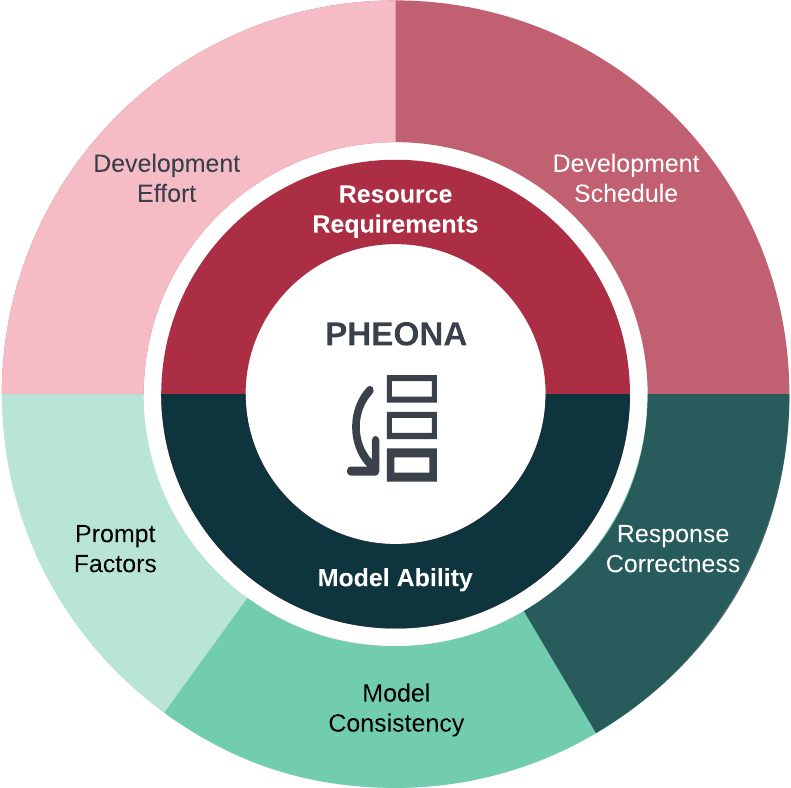}
 \caption{Overview of the components of \acrfull{pheona}.}
 \label{fig:pheona}
\end{figure}

We first discuss how individual components of \acrshort{pheona} (Figures \ref{fig:pheona} and \ref{fig:overview}) were constructed, including descriptions of the associated criteria. We then outline how we demonstrated \acrshort{pheona} on a specific phenotyping use case.

\subsection*{Development of \acrshort{pheona}}

Since \acrshort{llms} have not previously been applied to computational phenotyping tasks, the first component of \acrshort{pheona}, the resource requirements, was designed to determine when \acrshort{llm}-based methods may provide advantages over traditional phenotyping ones. The resource requirements outline considerations to estimate time, cost, and computational effort and should be considered before development begins. We based the resource requirements on components of \acrfull{cocomo} $2.0$, a cost estimation model for software development projects\cite{boehm_cost_1995}. Similar to \acrshort{cocomo}, we focused on attributes of the development effort (amount of work required) and development schedule (timeline of work required) when considering resource requirements for \acrshort{llm}-based projects. However, unlike \acrshort{cocomo}, we assumed an experienced research team so that development effort is primarily a function of existing hardware and software constraints. Table \ref{tab:resource_requirements} outlines individual attributes of the development effort and schedule for consideration.
\\
\\
\noindent
Following an assessment of whether \acrshort{llms} could be considered for a specific task, the second component of \acrshort{pheona}, model ability, provides criteria to evaluate \acrshort{llm}-specific processes (i.e., prompt engineering) and known drawbacks of \acrshort{llms} (i.e., hallucination and response instability)\cite{li_llms-as-judges_2024, zhuo_prosa_2024, patil_review_2024}. Therefore, model ability includes the following subcategories: 1) Prompt Factors; 2) Model Consistency; and 3) Response Correctness. Prompt factors evaluate the effort required for prompt engineering as well as the corresponding response times for each prompt. Model consistency factors evaluate the ability of the model to provide consistent answers, even if changes are made to the prompt, since \acrshort{llms} can be sensitive to inputs\cite{li_llms-as-judges_2024, zhuo_prosa_2024}. Additionally, automated response extraction from \acrshort{llm} responses is necessary for large volume requests and responses must be consistent to produce reliable results. Finally, response correctness factors were designed to evaluate overall performance of the model, including the presence of hallucinations. These criteria were most closely related to traditional phenotyping evaluation, where phenotyping performance is measured against ground truths.\cite{carrell_general_2024} Table \ref{tab:model_ability} details the model ability requirements. We note that when evaluating the model ability criteria, a random subset of the entire dataset may be necessary for evaluation, especially for large datasets without previously defined ground truths.

\subsection*{Phenotyping Use Case}

We applied \acrshort{pheona} to concept classification, a single development step for previously defined computable phenotypes for \acrfull{arf} respiratory support therapies. Although the phenotypes were discussed in-depth previously, we provide a brief overview to contextualize our work\cite{essay_rule-based_2020}. The phenotyping algorithm classified individuals based on the respiratory therapies received, specifically \acrfull{imv}, \acrfull{nippv}, and \acrfull{hfni} during an individual \acrfull{icu} stay. The respiratory therapies were identified by the presence of specific concepts,  such as ``\textit{Hi Flow NC}" and ``\textit{BiPAP/CPAP}", that indicated therapy use. Additionally, for \acrshort{imv}, presence of sedation and paralysis medications was required to verify patients underwent intubation. In the original study, manual review of the data elements was performed to determine which concepts were relevant to any of the respiratory therapies or specific medications. For our \acrshort{llm}-based classification, we engineered prompts to determine which concepts were relevant to at least one of the following categories: \acrshort{imv}, \acrshort{nippv}, \acrshort{hfni}, and medications used for sedation and paralysis.

\begin{figure}[H]
  \centering
    \includegraphics[width=0.9\textwidth]{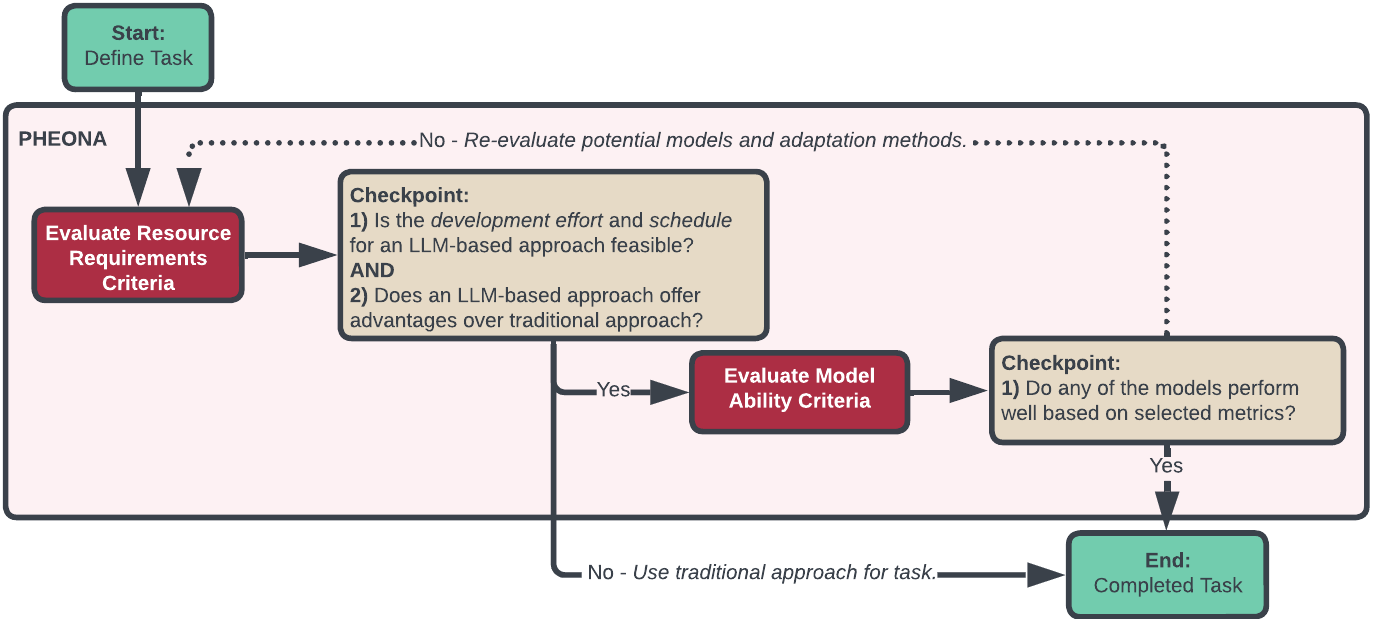}
 \caption{Overview of how to apply \acrfull{pheona} to a computational phenotyping task.}
 \label{fig:overview}
\end{figure}

\subsection*{Data Processing}

We used the \acrfull{eicu} dataset since concept ground truths existed prior to this study from development of the \acrshort{arf} respiratory support therapy phenotypes\cite{essay_rule-based_2020, pollard_eicu_2018}. \acrshort{ehr} data were processed to generate constructed data concepts using \acrshort{eicu} tables with timestamped data that likely contained information on respiratory therapies based on the table descriptions. Tables that did not include timestamped data but contained information on respiratory therapies (such as the \textit{apacheApsVar} table) or that were unlikely to contain descriptions of respiratory therapies (such as the \textit{vitalPeriodic} table) were excluded from further processing. There were $9$ total tables selected that were used to construct input concepts. Each constructed concept followed the format: \textit{Source: ``Table Name"; Concept = ``Concept Name and Details"}, where \textit{Table Name} was the name of the input table (e.g., \textit{Infusion Drug}), and \textit{Concept Name and Details} was a string concatenation of potentially relevant columns (see Table \ref{tab:processing}).

\subsection*{Prompt Engineering}

Prompts were developed and refined using a random sample of $100$ constructed concepts. Since best results on the sample were obtained when categorization of respiratory therapies (\acrshort{imv}, \acrshort{nippv}, and \acrshort{hfni}) was separate from categorization of sedation and paralysis medications, we developed two prompts, resulting in two \acrshort{llm} responses per constructed concept\cite{pungitore_prompts_2025}. The prompts were structured into multiple sections, including input, instructions, concept definitions, and output. Each constructed concept was injected into the input section. After prompt development, we required only the output section of the prompt for model ability evaluation, which included a series of questions and answers to guide the model into determining whether the constructed concept was relevant to either the respiratory therapies or medications of interest. The answer to the final question for each prompt was parsed using string methods to get the final response of ``YES" or ``NO" for whether the constructed concept was relevant. Since there were two prompts used, the final response was ``YES" if at least one of the responses was ``YES" and ``NO," otherwise.

\subsection*{Model Setup}

The models tested were identified from popular models available through Ollama, an open-source package that allows local hosting of open-source \acrshort{llms} \cite{noauthor_ollamaollama_2024}. Due to \acrfull{gpu} resource constraints, we selected the following instruction-tuned models for analysis: Mistral Small $24$ billion with Q$8.0$ quantization (model tag: 20ffe5db0161), Phi-$4$ $14$ billion with Q$8.0$ quantization (model tag: 310d366232f4), Gemma$2$ $27$ billion with Q$8.0$ quantization (model tag: dab5dca674db), and DeepSeek-r1 $32$ billion with Q$4$\_K\_M quantization (model tag: 38056bbcbb2d)\cite{noauthor_ollamaollama_2024}. Models were run on either a single Nvidia V$100$ $32$GB or two Nvidia P$100$ $16$GB \acrshort{gpus} depending on server availability. Temperature and top-p were set at $0.0$ and $0.99$, respectively, for all experiments.

\subsection*{Assessment of Resource Requirements}

The resource requirements were assessed by comparing manual review of the original $83{,}289$ concepts to automated \acrshort{llm} classification. In the original study, manual review of all concepts was performed by $3$ experts over an estimated $6$ months out of the total $18$ months required for phenotype development. Development effort and schedule for the \acrshort{llm}-based approach were measured during prompt engineering for the initial sample of $100$ constructed concepts.

\subsection*{Assessment of Model Ability}

To assess model ability, we evaluated all components of \acrshort{pheona} to fully demonstrate the framework and then selected the best model for concept classification based on response correctness, response latency, and response format accuracy. We prioritized response correctness to ensure accurate classification results. Response latency and response format accuracy were secondary evaluation metrics to ensure the model could quickly and consistently categorize all concepts. The time to minimal viable prompt was measured on the initial sample of $100$ constructed concepts and the remaining metrics were measured using another random sample of $100$ constructed concepts to avoid data leakage. We demonstrated the model ability criteria using data samples to mimic phenotype development where ground truths are unlikely to be available and it would be necessary to manually classify concepts prior to evaluation.
\\
\\
\noindent
For prompt factors, the time to minimum viable prompt was measured as the time to develop prompts for classification of the sample constructed concepts for each model. Model response latency was calculated by first summing the response latencies across the two \acrshort{llm} responses and then averaging the latency for the $100$ samples. Since we used internal environments for hosting the \acrshort{llms}, we did not consider testing at different times of the day. The response format accuracy was determined by counting the number of responses that were returned in the exact output format specified in the prompts\cite{pungitore_prompts_2025}. We tested response consistency and prompt stability by first randomly selecting $10$ constructed concepts from the second sample of $100$. For response consistency, we generated $10$ model results per constructed concept and evaluated how many times the model returned the same answer (either ``YES" or ``NO") for a total of $200$ responses per model. The prompt stability was tested by rearranging the respiratory therapy categorization prompt (since it was large and complex) and evaluating whether the model returned the same answer. The changes made to the prompt were the following: 1) Moved the instructions after the concept criteria; 2) Reversed the ordering of the concept-related questions; and 3) Reversed the ordering of the concepts within the prompt. Each change was independently tested $10$ times on each of the $10$ sampled constructed concepts for a total of $200$ responses per change.

For response correctness, the accuracy was measured using the \acrfull{roc} score. The quasi-accuracy was also measured with \acrshort{roc} after reclassifying any relevant false positives as true positives based on the prompt concept definitions after manual review\cite{pungitore_prompts_2025}. Hallucination frequency was determined by assessing the number of factually incorrect statements, such as incorrect acronym interpretations or incorrect association of medications to their functional group. The severity of each hallucination was evaluated as follows: Minor (little to no impact on the model reasoning or final decision); Major (impact on model reasoning but not the final decision, i.e., the right decision made for the wrong reason); Critical (impact on both model reasoning and final decision).

\section*{Results}

\subsection*{Assessment of Resource Requirements}

The resource requirements comparison for both methods is presented in Table \ref{tab:resource_requirements_demo}. From this comparison, we note that while the \acrshort{llm}-based methods required more specialized hardware and software, the time savings for classification was estimated to be significant when compared to manual review in the traditional approach, especially when considering that the actual concept classification is performed by the model. Therefore, we decided \acrshort{llm}-based methods would be appropriate to explore for concept classification.

\subsection*{Assessment of Model Ability}

For time to minimum viable prompt, early testing showed that a prompt developed on one model was effective for the remaining models and therefore, we did not estimate any differences in this metric between models. For the response correctness metrics, there were no hallucinations detected across any of the models. Additionally, all of the models achieved an \acrshort{roc} of $1.000$ for quasi-accuracy after reviewing the concepts identified as false positives. The false positives reassigned as true positives were \textit{Source = Nurse Charting; Concept = O2 Admin Device: BiPAP/CPAP} (all models) and \textit{Source = Nurse Charting; Concept = O2 Admin Device: nasal cannula} (Gemma and Mistral, only) since they were both relevant to the respiratory therapy concepts of \acrshort{nippv} and \acrshort{hfni}, respectively, as defined in the classification prompts\cite{pungitore_prompts_2025}. The results for the remaining metrics are presented in Table \ref{tab:model_ability_demo}. DeepSeek had the highest model response latency ($40.0$ seconds compared to $17.2$ seconds or faster) and lowest response format accuracy ($149/200$ compared to $177/200$ or better). Additionally, response format accuracy was lower for Gemma ($177/200$) when compared to Mistral ($197/200$) or Phi ($194/200$), indicating Gemma was less suitable for concept classification. Since Mistral and Phi had similar metrics for model response latency and response format accuracy, we would select either of these models for categorization of the entire constructed concept dataset.

\section*{Discussion}

In this study, we developed \acrshort{pheona}, an evaluation framework specifically for applying \acrshort{llms} to computational phenotyping. In addition to outlining components of \acrshort{pheona}, we demonstrated how to apply it to concept classification, a single task within a broader phenotyping pipeline for \acrshort{arf} respiratory support therapies. As \acrshort{llm}-based methods continue to evolve, we encourage corresponding modifications to \acrshort{pheona} to support these changes.

\subsection*{Evaluation Framework}

The primary contribution of this study was development of \acrshort{pheona} to support advancement of computational phenotyping methods using \acrshort{llms}. Despite recent improvements, computational phenotyping remains complicated and costly\cite{shang_making_2019, tekumalla_towards_2024} with many of the desiderata for computational phenotyping identified over a decade ago still relevant today\cite{hripcsak_next-generation_2013, wen_impact_2023}. Foundation models like \acrshort{llms} are the first widely available tools with the capability to comprehend and produce human-readable text for a fraction of the development time and thus, they are a natural next step for computational phenotyping. With \acrshort{pheona}, we provided guidelines to evaluate \acrshort{llm} integration into phenotyping pipelines. The resource requirements component was designed to assess whether \acrshort{llms} are appropriate for the task of interest since not every task could or should use \acrshort{llms} based on available resources. The model ability component was then designed to determine if available \acrshort{llms} could accomplish the task of interest. To simplify development and avoid additional costs, we suggest starting with smaller parameter \acrshort{llms} and consider larger models if performance is low. There are other methods to consider, such as fine-tuning\cite{li_llms-as-judges_2024}, but these may not provide an advantage over traditional methods and should be evaluated on a case-by-case basis. In these scenarios, the application of \acrshort{pheona} would look similar to Figure \ref{fig:overview}, where the resource requirements and model ability are cycled through to find the optimal method.

\subsection*{Phenotyping Use Case}

Another contribution was an initial assessment of concept classification for \acrshort{arf} respiratory support therapies. From our model ability criteria evaluation, we concluded either Mistral or Phi could effectively perform concept classification. Since all of the models had high response correctness scores, we only focused on secondary metrics (model response latency and response format accuracy) for model evaluation. With respect to these metrics, DeepSeek performed marginally worse than the other models; however, it was also the only model to incorporate reasoning into its response, which makes the responses less consistent and more lengthy but potentially more valuable for complex tasks. These results demonstrated several advantages of \acrshort{llm}-based approaches specifically for classification tasks. First, these results showed high accuracy can be achieved without extensive feature engineering or model training. Second, our methods incurred no additional hardware costs, indicating that \acrshort{llm}-based methods do not need expensive workflows to be effective for classification. Finally, our results were achieved with minimal data processing and relatively fast latency, suggesting potential further adaptation of \acrshort{llms} to time-sensitive or data-intensive clinical workflows.

\subsection*{Study Limitations}

There are known issues with using \acrshort{llms}, including the ability to produce hallucinations and reduced performance around specialized, domain-specific knowledge\cite{li_llms-as-judges_2024, patil_review_2024}. While the model ability components of \acrshort{pheona} help highlight where \acrshort{llms} may fail, there may be cases where \acrshort{llms} cannot accomplish the task of interest and traditional phenotyping methods must be used instead. Additionally, the current version of \acrshort{pheona} requires manual review of responses to measure each criterion, despite the known drawback of manual review being time-intensive. Although we used a random sample of constructed concepts to make manual review more feasible, it is possible performance on the entire dataset is not representative of performance on the random sample. Finally, for our use case, we only considered the binary classification task of ``YES/NO" for whether each constructed concept was relevant to the topics of interest although multi-class classification may be more useful for downstream processes. However, since we did not observe any inaccurate responses when reviewing responses to individual classification steps within our prompt, we should be able to convert from binary to multi-class classification using prompt updates. For example, we could ask the model to return the most relevant therapy (\acrshort{imv}, \acrshort{nippv}, or \acrshort{hfni}) rather than “YES/NO.”

\subsection*{Future Directions}

Since one of the limitations of this study was the reliance on manual review for evaluation, there are opportunities to develop automated \acrshort{llm} agents for computational phenotyping and update \acrshort{pheona} to incorporate automated evaluation processes\cite{qiu_llm-based_2024}. Furthermore, while we demonstrated \acrshort{pheona} on a single task, the framework can be generalized to adapt \acrshort{llm}-based methods to entire computational phenotyping pipelines rather than simply adapting \acrshort{llms} to individual phenotyping tasks. Finally, each of the criteria in \acrshort{pheona} can be expanded upon to include a more standardized evaluation approach. For example, a framework developed to measure \acrshort{llm} sensitivity to general inputs can be adapted to the prompt stability component of \acrshort{pheona}\cite{zhuo_prosa_2024}.

\section*{Conclusions}

We developed \acrshort{pheona}, an evaluation framework for applying \acrshort{llm}-based methods to computational phenotyping tasks. This framework helps assess if \acrshort{llms} are feasible and if so, which model is best for each task. We demonstrated \acrshort{pheona} on an example use case to promote further research into improving computational phenotyping.

\cleardoublepage

\begin{table}[ht]
	\vspace{10pt}
	\caption{Evaluation criteria for resource requirements, specifically attributes of development effort and development schedule. These criteria were designed to compare the necessary resources for \acrshort{llm}-based approaches to traditional phenotyping approaches.}
	\label{tab:resource_requirements}
	{
		\renewcommand{\arraystretch}{1.0}
		\begin{tabular}{>{\raggedright\arraybackslash}p{0.15\linewidth}
				>{\raggedright\arraybackslash}p{0.37\linewidth}>{\raggedright\arraybackslash}p{0.40\linewidth}
			}
			
			\hline
			\multicolumn{3}{l}{\textbf{Development Effort}} \\ 
			\hline
			\textbf{Criterion} & \textbf{Description} & \textbf{How to Measure} \\ 
			\hline
			
			Model Host Environments
			&
			Availability of external and internal environments for hosting models and artifacts based on project budget, expected timeline, \acrfull{phi} restrictions, and/or available compute environments.
			&
			Consider each restriction from most to least significant and use to determine the list of feasible environments to host models. For example, use of \acrshort{phi} may limit access to externally hosted models if privacy safeguards cannot be guaranteed through standard end-user license agreements.
			\\
			Hardware Requirements 
			&
			Minimum hardware requirements required to run model(s) of interest if considering internal environments. May be re-evaluated to accommodate any inference optimization strategies implemented\cite{chitty-venkata_survey_2023}.
			&
			Assess the maximum computational resources in each of the available host environments, focusing on \acrfull{gpu} resources in particular. Assess whether maximum resources will be available throughout the project or if a smaller model size is desired to ensure consistent access.
			\\
			Software Availability 
			&
			For externally hosted models, consideration of the availability of software to \textit{interact} with these models. For internally hosted models, consideration of the availability and complexity of software to \textit{host and interact} with selected models.
			&
			Assess the software requirements to interact with models in each of the available host environments and determine time to setup and develop with available software.
			\\
			\hline
			\multicolumn{3}{l}{\textbf{Development Schedule}} \\ 
			\hline
			\textbf{Criterion} & \textbf{Description} & \textbf{How to Measure} \\ 
			\hline
			
			Time for Pipeline Development
			&
			The amount of time required to implement the desired phenotyping methods, including development and evaluation time.
			&
			Estimate the required development time to complete each of the phenotyping steps. For \acrshort{llm}-based methods these steps would likely encompass model setup, data processing, prompt engineering, and response evaluation.
			\\
			
			Time for Manual Review
			&
			The amount of time required for manual review of responses based on the phenotyping task and overall pipeline.
			&
			Based on the steps in the phenotyping pipeline that would require manual review, estimate total review time based on quantity of data and any initial results.
			\\
			
			Phenotype Runtime
			&
			The amount of time required to perform all included phenotyping tasks on one record, such as data processing time, model runtime, and/or any \acrfull{api} response times.
			&
			Estimate runtimes based on size of data and complexity of phenotyping process. For model \acrshort{api} calls, use example data and prompts to estimate total phenotyping time. If using an externally hosted model or other external resources, consider testing \acrshort{apis} at different times of the day to test latency variations.
			\\
			\hline
			
		\end{tabular}
	}
\end{table}

\cleardoublepage

\begin{table}[ht]
	\vspace{10pt}
	\caption{Evaluation criteria for each subcategory under model ability. These criteria were designed to determine the model that achieves optimal results on specific phenotyping tasks.}
	\label{tab:model_ability}
	{
		\renewcommand{\arraystretch}{1.15}
		\begin{tabular}{>{\raggedright\arraybackslash}p{0.15\linewidth}
				>{\raggedright\arraybackslash}p{0.30\linewidth}>{\raggedright\arraybackslash}p{0.48\linewidth}
			}
			
			\hline
			\multicolumn{3}{l}{\textbf{Prompt Factors}} \\ 
			\hline
			\textbf{Criterion} & \textbf{Description} & \textbf{How to Measure} \\ 
			\hline
			Time to Minimum Viable Prompt 
			&
			The amount of time to engineer prompt with acceptable results.
			&
			Estimate prompt engineering time after testing different prompts and prompt structures (such as few-shot learning \cite{liu_pre-train_2023}), considering how adaptable each model is to increasing prompt complexity.
			\\
			Model Response Latency
			&
			The time to return responses for each prompt.
			&
			Measure the average amount of time to receive responses from the \acrshort{llm} for each prompt of interest. If using an externally hosted model, consider testing at different times of the day to assess latency variations.
			\\
			\hline
			\multicolumn{3}{l}{\textbf{Model Consistency}} \\ 
			\hline
			\textbf{Criterion} & \textbf{Description} & \textbf{How to Measure} \\ 
			\hline
			Response Format Accuracy &
			The accuracy of correctly formatted responses based on the requested format in the prompt. &
			From the requested structure in the prompt, determine how many responses were returned in this format with no deviations.
			\\ 
			Response Consistency &
			The consistency of responses with multiple model runs.  &
			Run each model multiple times with the same prompt on the same data and record how frequently model returns same answer.
			\\ 
			Prompt Stability &
			The ability of the prompts to return consistent responses even if there are perturbations in the prompts.
			&
			Make changes to the prompt and re-evaluate relevant criteria, such as response format accuracy and classification accuracy. Example perturbations include reordering prompt sections and removing any text emphasis.
			\\
			\hline
			\multicolumn{3}{l}{\textbf{Response Correctness}} \\ 
			\hline
			\textbf{Criterion} & \textbf{Description} & \textbf{How to Measure} \\ 
			\hline
			Accuracy & 
			How accurate the model is at classifying or phenotyping based on a ground truth. Based on task requirements, this may also include assessing the accuracy of responses with respect to the reference text or provided context. & 
			Classification accuracy can be measured using common classification metrics like  accuracy, precision, recall, and \acrfull{roc}. Accuracy against reference text can be performed using N-gram analysis, \acrshort{bleu}, or \acrshort{rouge} while accuracy against provided context can be performed also by N-gram analysis and additionally \acrshort{rouge}-C.\cite{sai_survey_2022}
			\\
			Quasi-Accuracy & 
			Classification accuracy after marking relevant or semantically similar false positives as true positives to avoid penalizing \acrshort{llm} when unnecessary. &
			Review responses classified as False Positive and determine if either a) the original text is semantically relevant to the classification task of interest or b) there is evidence that the ground truth is incorrect. Recalculate classification metrics after reassigning these cases to True Positives.
			\\
			Hallucination Frequency and Severity &
			Determine the frequency and corresponding severity of hallucinations in \acrshort{llm} responses. &
			Assess frequency of factually incorrect statements in responses. For each hallucination, assess severity based on, for example, whether the outcome was affected, severe biases were introduced, or the response detracted from the overall goal of the prompt.
			\\
			\hline
			
		\end{tabular}
	}
\end{table}

\begin{table}[ht]
	\vspace{10pt}
	\caption{Constructed concept pattern for each selected table in the \acrfull{eicu} database\cite{pollard_eicu_2018}. Italicized text was replaced with specific values from the relevant column in each table.}
	\label{tab:processing}
	{
		\renewcommand{\arraystretch}{1.0}
		\begin{tabular}{>{\raggedright\arraybackslash}p{0.18\linewidth} | >{\raggedright\arraybackslash}p{0.78\linewidth}} 
			
			\hline
			\textbf{Table Name} & \textbf{Constructed Concept Pattern} \\ 
			\hline

			Care Plan General & Source = Care Plan General; Concept = \textit{cplgroup}: \textit{cplitemvalue} \\
			
			Infusion Drug & Source = Infusion Drug; Concept = \textit{drugname} \\ 
			
			Medication & Source = Medication; Concept = \textit{drugname} \\ 
			
			Note & Source = Note; Concept = \textit{notevalue}: \textit{notetext} \\ 
			
			Nurse Care & Source = Nurse Care; Concept = \textit{cellattributevalue} \\ 
			
			Nurse Charting & Source = Nurse Charting; Concept = \textit{nursingchartcelltypevalname}: \textit{nursingchartvalue} \\ 
			
			Respiratory Care & Source = Respiratory Care; Concept = \textit{airwaytype} \\ 
			
			Respiratory Charting & Source = Respiratory Charting; Concept =  \textit{respcharttypecat}: \textit{respchartvaluelabel}: \textit{respchartvalue} \\ 
			
			Treatment & Source = Treatment; Concept = \textit{treatmentstring} \\ 
			
			\hline
			
		\end{tabular}
	}
\end{table}


\cleardoublepage

\begin{table}[ht]
	\caption{\fontsize{10}{10} Demonstration of the resource requirements to decide between an \acrshort{llm}-based or a traditional phenotyping approach for the task of concept classification in a computable phenotype for \acrfull{arf} respiratory support therapies.}
	\label{tab:resource_requirements_demo}
	
	{
		\renewcommand{\arraystretch}{1.0}
		\begin{tabular}{>{\raggedright\arraybackslash}p{0.13\linewidth} 
				>{\raggedright\arraybackslash}p{0.25\linewidth} 
				>{\raggedright\arraybackslash}p{0.29\linewidth} 
				>{\raggedright\arraybackslash}p{0.23\linewidth}
			}
			
			\hline
			\multicolumn{4}{l}{\textbf{Development Effort}} \\ 
			\hline
			\textbf{Criterion} & 
			\textbf{Traditional Approach} & 
			\textbf{\acrshort{llm}-based Approach} & 
			\textbf{Verdict} \\ 
			\hline
			
			Model Host Environments
			&
			Due to \acrshort{phi} concerns, only internal environments were available for use.
			&
			Due to \acrshort{phi} concerns, only internal environments were available for use. Environments with \acrshort{gpu} resources were available but restricted based on demand.
			&
			Traditional approach was superior due to ability to run model on multiple internal environments rather than being restricted to those with \acrshort{gpu} access.
			\\ 
			
			Hardware Requirements
			&
			No additional computational resources or cost expected.
			&
			Access to high performance research servers was limited based on demand but incurred no additional cost.
			&
			Traditional approach was marginally superior due to limits on \acrshort{gpu} resources although neither approach incurred additional costs.
			\\
			
			Software Requirements
			&
			No additional software requirements specific to phenotyping required.
			&
			Software specific to interacting with \acrshort{llm} models was required and troubleshooting was required to adapt software to high performance compute environment.
			&
			Traditional approach was superior due to extra setup time for \acrshort{llm} software.
			\\
			
			\hline
			\multicolumn{4}{l}{\textbf{Development Schedule}} \\ 
			\hline
			\textbf{Criterion} & 
			\textbf{Traditional Approach} & 
			\textbf{\acrshort{llm}-based Approach} & 
			\textbf{Verdict} \\ 
			\hline
			
			Time for Pipeline Development
			&
			An estimated $6$ months of the total $18$ months for \acrshort{arf} phenotype development were required for manual review of concepts.
			&
			A total of $2$ months was required for prompt engineering on the initial sample of concepts. An estimated additional month would be required to gather all concept responses for all models based on sample latencies with models running in parallel.
			&
			\acrshort{llm}-based method was superior.
			\\
			
			Time for Manual Review
			&
			An estimated $6$ months of the total $18$ months for \acrshort{arf} phenotype development were required for manual review of concepts.
			&
			A week was required for clinical experts to collate information on the concepts of interest to be used for prompt engineering and an additional day was required to manually review model ability results.
			&
			\acrshort{llm}-based method was superior.
			\\
			
			Phenotype Runtime
			&
			There were no artifacts specifically for concept classification.
			&
			An estimated $30$ seconds was required to classify each concept as relevant for the \acrshort{arf} respiratory support phenotypes.
			&
			Not Applicable, since the traditional phenotyping methods relied only on manual review for concept classification.
			
			\\
			\hline
			
		\end{tabular}
	}
\end{table}

\cleardoublepage


\begin{table}[ht]
	\vspace{10pt}
	\caption{Demonstration of the model ability criteria to determine best \acrshort{llm} for concept classification. When included, the denominators indicate the number of \acrfull{llm} responses that were evaluated for each metric based on the specific experiment implemented.}
	\label{tab:model_ability_demo}
	{
		\renewcommand{\arraystretch}{1.0}
		\begin{tabular}{>{\raggedright\arraybackslash}p{0.28\linewidth}
				>{\raggedright\arraybackslash}p{0.15\linewidth}>{\raggedright\arraybackslash}p{0.15\linewidth}>{\raggedright\arraybackslash}p{0.15\linewidth}>{\raggedright\arraybackslash}p{0.15\linewidth}
			}
			
			\hline
			\multicolumn{5}{l}{\textbf{Prompt Factors}} \\ 
			\hline
			\textbf{Criterion} & \textbf{DeepSeek} & \textbf{Gemma} & \textbf{Mistral} & \textbf{Phi} \\ 
			\hline
			
			Model Response Latency &
			$40.0$ seconds
			&
			$17.2$ seconds
			&
			$14.8$ seconds
			&
			$11.3$ seconds
			\\ 
			\hline
			\multicolumn{5}{l}{\textbf{Model Consistency}} \\ 
			\hline
			\textbf{Criterion} & \textbf{DeepSeek} & \textbf{Gemma} & \textbf{Mistral} & \textbf{Phi} \\ 
			\hline
			
			Response Format Accuracy &
			$149/200\;(74.5\%)$
			&
			$177/200\;(88.5\%)$
			&
			$197/200\;(98.5\%)$
			&
			$194/200\;(97.0\%)$
			\\ 
			Response Consistency &
			$197/200\;(98.5\%)$
			&
			$200/200\;(100\%)$
			&
			$200/200\;(100\%)$
			&
			$200/200\;(100\%)$
			\\
			
			Prompt Stability &
			&
			&
			&
			\\ 
			\;\;\;\;\;Instructions moved &
			$199/200\;(99.5\%)$
			&
			$200/200\;(100\%)$
			&
			$190/200\;(95.0\%)$
			&
			$200/200\;(100\%)$
			\\ 
			\;\;\;\;\;Questions reversed &
			$200/200\;(100\%)$
			&
			$200/200\;(100\%)$
			&
			$200/200\;(100\%)$
			&
			$191/200\;(95.5\%)$
			\\ 
			
			\;\;\;\;\;Concept order reversed &
			$199/200\;(99.5\%)$
			&
			$200/200\;(100\%)$
			&
			$200/200\;(100\%)$
			&
			$200/200\;(100\%)$
			\\ 
			\hline
			\multicolumn{5}{l}{\textbf{Response Correctness}} \\ 
			\hline
			\textbf{Criterion} & \textbf{DeepSeek} & \textbf{Gemma} & \textbf{Mistral} & \textbf{Phi} \\ 
			\hline
			Accuracy (\acrshort{roc}$^a$) & 
			$0.995$
			&
			$0.990$
			&
			$0.990$
			&
			$0.995$
			\\
			\hline
			\multicolumn{5}{l}{$^a$ \acrshort{roc}: \acrlong{roc} curve.}
		\end{tabular}
	}
\end{table}

\cleardoublepage
\makeatletter
\renewcommand{\@biblabel}[1]{\hfill #1.}
\makeatother

\printbibliography[heading=subbibliography]

\end{document}